\documentclass[conference,letterpaper]{IEEEtran}
\usepackage{amsmath}
\usepackage{amssymb}
\usepackage{graphicx}
\usepackage{epstopdf}
\usepackage{multirow}
\usepackage{booktabs}
\usepackage{placeins}
\usepackage{dblfloatfix}

\newcommand{\tablebody}{%
    \footnotesize
    \setlength{\tabcolsep}{4pt}%
    \renewcommand{\arraystretch}{1.1}%
}

\title{\vspace*{0.25in}\fontsize{15pt}{18pt}\selectfont
 TAO-DA: Towards Autonomous Operation---A Dual-Arm Vision--Language--Action Model for Coordinated Manipulation}

\author{
\IEEEauthorblockN{
Yongsheng Zhao,
Han Gao,
Baoping Cheng,
Jingyao Tang,
Dian Zhou,\\
Deng Liang,
Ji Ge,
Xuanzhang Wen,
Lei Zhao,
Ye Wang
}
\IEEEauthorblockA{
China Mobile (Hangzhou) Information Technology Co., Ltd., Hangzhou, China\\
Corresponding author: Baoping Cheng, chengbaoping@cmhi.chinamobile.com
}
}

\begin{document}
\maketitle

\begin{abstract}
Vision–Language–Action (VLA) models provide a unified framework for grounding high-level semantic information into low-level robot actions, enabling scalable robotic manipulation across diverse tasks. However, existing VLA models lack explicit mechanisms to disentangle the states and intents of the two arms, leading to unintended cross-arm interference that degrades task execution success. To address this issue, we propose a symmetric Dual‑Arm Expert (DAE) architecture built upon a shared Vision–Language Model (VLM) backbone with decoupled, arm‑specific expert towers. Expert selection is carried out through a two-stage dual-arm intent routing scheme, in which experts are routed either by explicit language instructions in the first stage or by implicit visual semantics in the second stage. Moreover, we introduce a lightweight task progress prediction module that leverages cross-attention between the pre-chunk temporal features and semantic representations of proprioceptive and visual observations to accurately estimate frame-wise task completion progress. This module facilitates task progress synchronization to support coordinated scheduling for collaborative multi‐robot tasks. Experimental results demonstrate the effectiveness of our model in dual-arm intent routing and the disentanglement of cross-arm interference, and further provide preliminary evidence of emergent skill generalization from single- to dual-arm tasks (as well as the reverse), together with cross-arm motion-domain skill transfer.
\end{abstract}


\section{INTRODUCTION}

Vision--Language--Action (VLA) models combine language, visual observations, and robot states to generate manipulation actions, providing a language-conditioned interface for control~\cite{ma2024survey}. Recent systems use autoregressive, diffusion-based, or flow-matching action generators for structured manipulation tasks~\cite{kim2024openvla,black2024pi0,bjorck2025gr00t,chi2025diffusion}.

A common bimanual VLA design represents both arms in a unified action generator. This simplifies the action interface but may couple the executing arm to irrelevant state and visual changes from the other arm. Such coupling matters when only one arm is required, when the arms execute sequential subtasks, or when demonstrations assign asymmetric roles. A dual-arm policy should therefore retain shared task semantics while selectively controlling arm-level action generation.

Based on this observation, we propose TAO-DA, a dual-arm VLA framework built around two principles: shared multimodal representations for task-level reasoning and arm-specific action towers for low-level control. DAIR determines the active-arm set from language and visual context, and DAE generates actions through the selected arm-specific pathways. We evaluate this factorization against unified and weakly decoupled alternatives.

The main contributions are as follows:

\begin{itemize}
    \item We formulate inactive-arm interference as an action-generation problem and study three DAE variants spanning unified, mask-based, and independent arm-level processing.
    \item We introduce Dual-Arm Intent Routing (DAIR), which combines rule-based language parsing with learned multimodal classification to predict left-only, right-only, or dual-arm execution.
    \item We evaluate the architectures on real-robot tasks under controlled inactive-arm disturbances, together with routing ablations under instructions that do not explicitly identify the active-arm set and complementary studies of arm-wise skill reuse.
\end{itemize}

\begin{figure*}[tb]
    \centering
    \includegraphics[width=.95\linewidth]{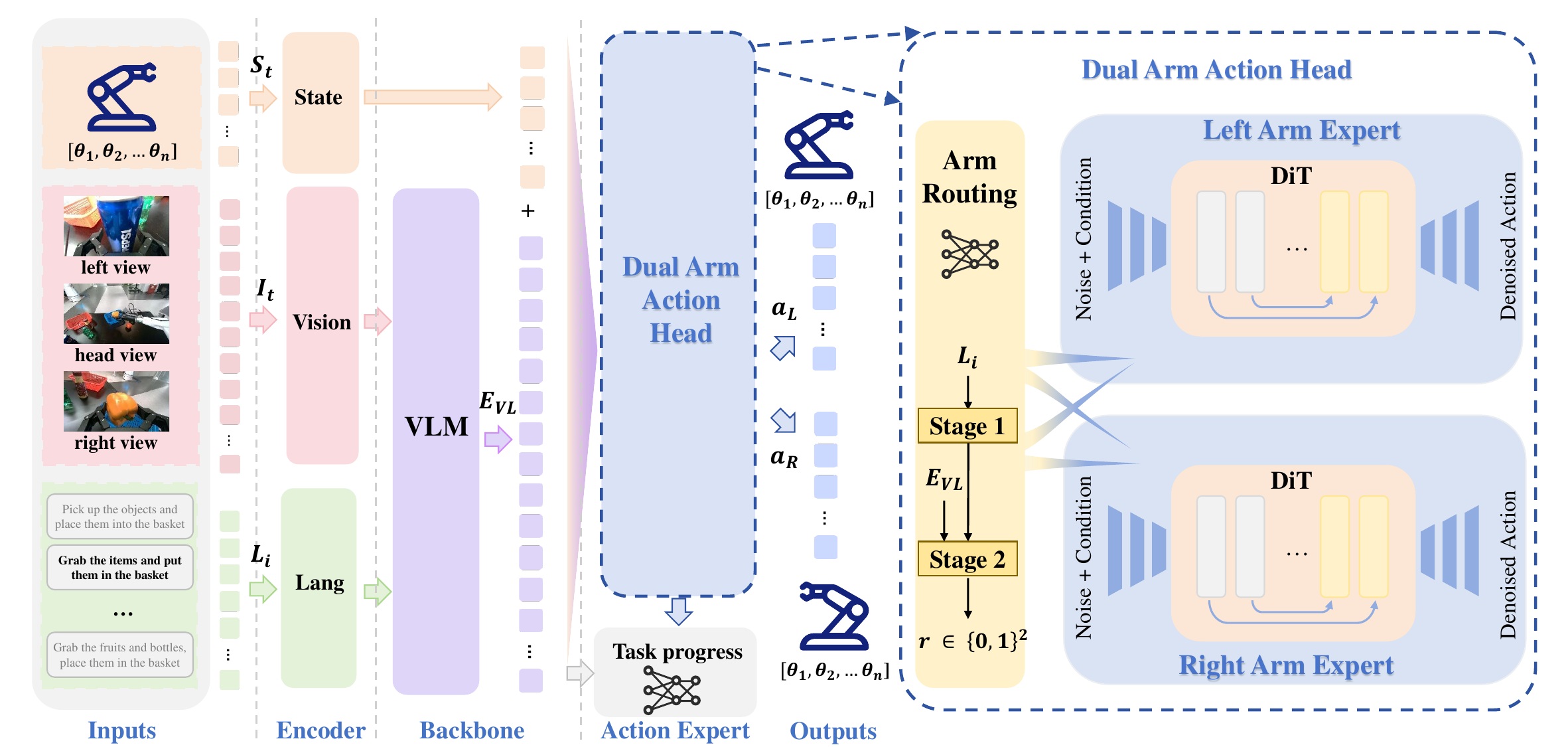}
    \caption{Overview of TAO-DA. SigLIP encodes the multi-view observations, and Eagle-2.5-VL fuses visual and language tokens into the context used by DAIR. After routing, inactive end-effector-view tokens are masked at the VLM input when constructing the action-conditioning context. DAE then generates actions through the selected arm-specific pathways, trained with flow matching.}
    \label{fig:architecture}
\end{figure*}

\section{RELATED WORK}

\subsection{Vision--Language--Action Models}
VLA models ground multimodal observations in robot actions. Autoregressive methods such as RT-1, RT-2, and OpenVLA discretize actions for sequence modeling~\cite{brohan2022rt,zitkovich2023rt,kim2024openvla}. Diffusion-based policies model continuous action distributions, while flow-matching methods such as $\pi_0$ and GR00T use continuous generative action models~\cite{chi2025diffusion,black2024pi0,bjorck2025gr00t}. These methods commonly use a unified action representation. TAO-DA instead studies whether the action generator should be factorized by arm.

\subsection{Structured Bimanual Manipulation}
Bimanual policies have explored task decomposition, visual masking, one-shot demonstrations, and transfer from single-arm policies. AnyBimanual uses skill management and visual alignment to coordinate unimanual primitives~\cite{lu2025anybimanual}; EgoVLA and VLBiMan learn bimanual behaviors from human or demonstration-based priors~\cite{yang2025egovla,zhou2025vlbiman}; TwinVLA reuses single-arm policies with additional coordination mechanisms~\cite{im2025twinvla}. Recent work introduces more explicit structure into bimanual VLA policies. PEAfowl strengthens geometry-aware multi-view fusion~\cite{fan2026peafowl}, while
\emph{See Selectively, Act Adaptively} and Co-VLA structure action generation through interaction-aware experts or shared and arm-specific coordination components~\cite{choi2026selectively,wang2026covla}. A complementary approach improves robustness through training-time modality masking~\cite{cheng2026robust}.

For compositional skill reuse, SkillVLA recombines learned single-arm skills across new left--right pairings~\cite{zhai2026skillvla}, whereas MA-VLA assigns mid-level atomic prompts to individual arms to support unseen collaboration patterns~\cite{zhang2026mavla}. These approaches motivate explicit role assignment and selective information flow. TAO-DA instead focuses on active-arm-set inference and disturbance isolation: DAIR selects left-only, right-only, or dual-arm execution, while Dual-Tower separates arm-level action-generation parameters.

\section{METHOD}

\subsection{Overview}

We propose TAO-DA (Towards Autonomous Operation with a Dual-Arm VLA), as illustrated in Fig.~\ref{fig:architecture}. Unlike approaches that model bimanual manipulation in a unified action space, TAO-DA retains a shared multimodal task representation while separating arm-specific action generation.

TAO-DA takes multi-view visual observations, a language instruction, and proprioceptive states as input. The observations comprise one egocentric-view image and two optional end-effector-view images captured by cameras mounted on the left and right arms. Each image is encoded by the same pretrained SigLIP vision encoder~\cite{zhaiSigmoidLossLanguage2023}. The resulting visual tokens are concatenated with language tokens and processed by the frozen Eagle-2.5-VL backbone to extract fused representations.

Let $\mathbf{E}_{\mathrm{VL}}\in\mathbb{R}^{N\times D}$ denote the unmasked multimodal representation used by DAIR. After routing, the action-conditioning representation is constructed with the inactive end-effector-view tokens masked at the VLM input. Proprioceptive features are encoded separately, while arm-specific states and action tokens are processed by their corresponding action pathways. During training, the instruction for each task is randomly sampled from a pool of semantically equivalent expressions.

\subsection{DAIR: Dual-Arm Intent Routing Guided by Language and Vision}

\subsubsection{Intent Representation}
We encode the active-arm set as the binary vector
\[
\mathbf{r}=[r_L,r_R]\in\{0,1\}^2,
\]
where $r_L$ and $r_R$ indicate whether the left and right arms are active. The valid classes are left-only, right-only, and dual-arm:
\[
\phi(0)=[1,0],\qquad
\phi(1)=[0,1],\qquad
\phi(2)=[1,1].
\]
We train the routing classifier as a three-class problem with cross-entropy loss:
\[
\mathcal{L}_{\mathrm{route}}=-\log p_c,
\]
where $c\in\{0,1,2\}$ is the target class and $p_c$ is the predicted probability of that class. The three-class interface prevents the classifier from producing the invalid no-arm vector $[0,0]$ during normal execution.

\subsubsection{Two-stage Arm Intent Inference}
\paragraph{Stage~1: Rule-Based Language Intent Parsing}
DAIR first applies a rule parser to the language instruction. The parser recognizes explicit arm-related expressions, such as ``left arm,'' ``right arm,'' ``both arms,'' ``left hand,'' and ``right hand,'' and maps them to the corresponding active-arm set $\hat{\mathbf{r}}_{\mathrm{rule}}$. When expressions referring to both arms are present, the dual-arm intent takes precedence and the parser returns $\hat{\mathbf{r}}_{\mathrm{rule}}=[1,1]$.

The rule parser does not use spatial expressions to infer arm intent. If an instruction contains no explicit arm-related expression, the parser returns the internal no-match code $[0,0]$, which indicates that the learned multimodal stage should be activated rather than a valid no-arm intent.

\paragraph{Stage~2: Model-Based Multimodal Intent Prediction}
In the second stage, the multimodal features are normalized and pooled using a learnable query:
\[
\begin{aligned}
\mathbf{E}_{\mathrm{norm}}
&=\operatorname{LayerNorm}(\mathbf{E}_{\mathrm{VL}}),\\
\mathbf{f}
&=\operatorname{MHA}(\mathbf{q},\mathbf{E}_{\mathrm{norm}},\mathbf{E}_{\mathrm{norm}}).
\end{aligned}
\]
An MLP produces three logits, which are converted to class probabilities $p_i$; the predicted class is then mapped to the binary interface:
\[
\hat{c}=\operatorname*{arg\,max}_{i}p_i,
\qquad
\hat{\mathbf{r}}_{\mathrm{model}}=\phi(\hat{c}).
\]

During inference, a previously selected intent can be held after the task has progressed sufficiently. Given the predicted progress $\hat{p}_t$, the threshold $\tau_{\mathrm{hold}}$, and the previous intent $\hat{\mathbf{r}}_{t-1}$, the routing rule is
\[
\hat{\mathbf{r}}_t=
\begin{cases}
\hat{\mathbf{r}}_{\mathrm{rule}},
& \hat{\mathbf{r}}_{\mathrm{rule}}\ne[0,0],\\
\hat{\mathbf{r}}_{\mathrm{model}},
& \substack{\hat{\mathbf{r}}_{\mathrm{rule}}=[0,0],\\
\hat{\mathbf{r}}_{t-1}\ \text{unavailable}},\\
\hat{\mathbf{r}}_{\mathrm{model}},
& \substack{\hat{\mathbf{r}}_{\mathrm{rule}}=[0,0],\\
\hat{p}_t\le\tau_{\mathrm{hold}}},\\
\hat{\mathbf{r}}_{t-1},
& \text{otherwise}.
\end{cases}
\]
This progress-based hold mechanism stabilizes routing decisions; it is distinct from the deployment-level task scheduler.

\subsection{DAE: Dual-Arm Expert Module}

\begin{figure}[tb]
    \centering
    \includegraphics[width=\linewidth]{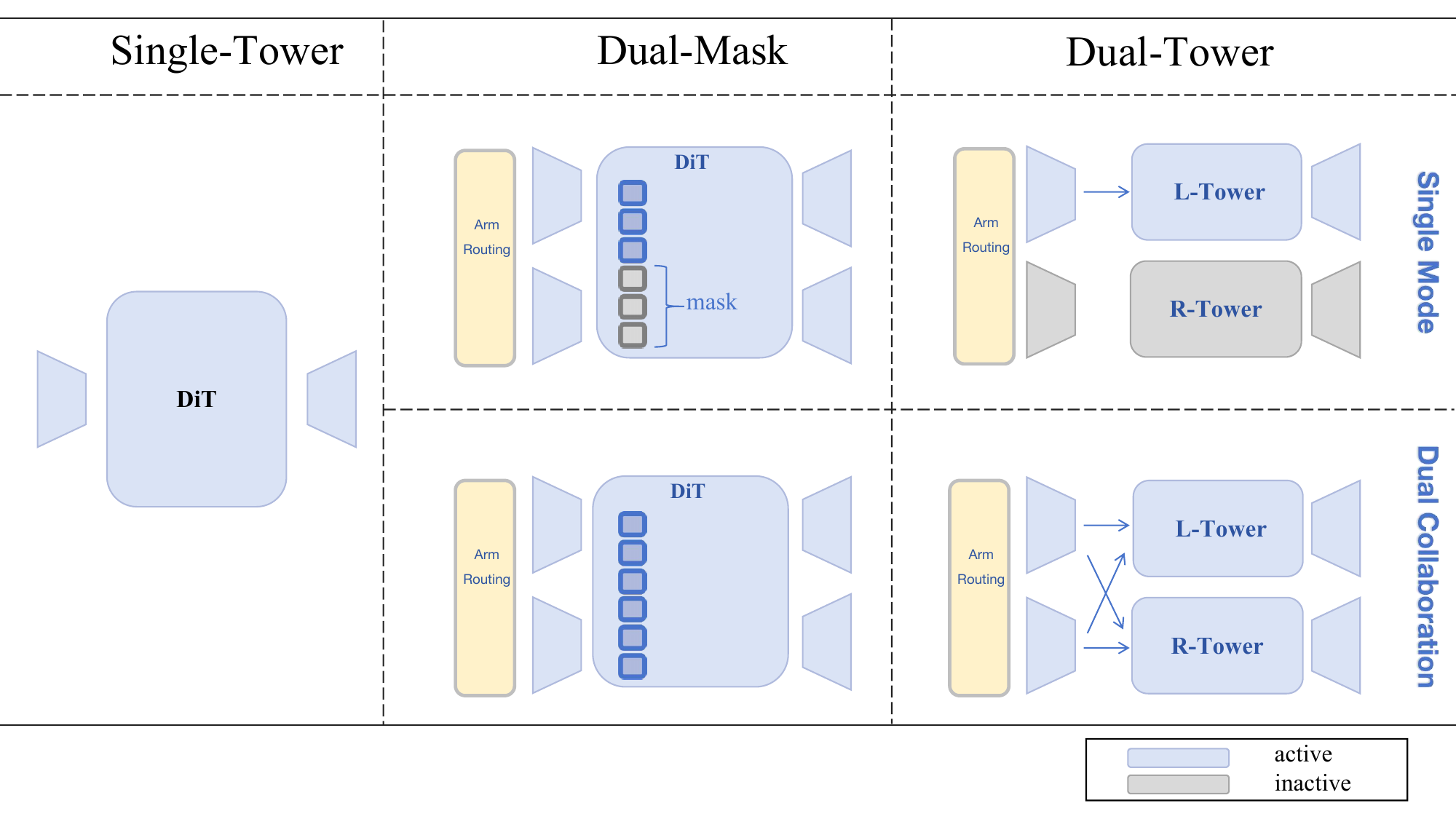}
    \caption{Architectural variants of DAE. Single-Tower uses a unified state and action space, Dual-Mask introduces weak decoupling through selective token masking in a shared DiT, and Dual-Tower provides strong decoupling through two parameter-independent DiT branches. All variants support single-arm and dual-arm execution.}
    \label{fig:compare3}
\end{figure}

We compare three DAE variants spanning unified to fully separated arm-level processing, as shown in Fig.~\ref{fig:compare3}. Single-Tower concatenates both arms' state and action tokens in one DiT; Dual-Mask retains the shared DiT but restricts token interactions; and Dual-Tower separates the branches and their parameters. All variants use the same multimodal context and flow-matching objective. Because Dual-Tower duplicates the arm-level components, this is an architectural comparison rather than a strictly parameter-matched attribution.

Unless otherwise specified, TAO-DA refers to the complete model comprising DAIR, the task-progress module, and the Dual-Tower DAE variant.

\subsubsection{Dual-Mask: Weak Decoupling via Token Masking}
\paragraph{Arm-Specific Token Masking}
Dual-Mask uses a shared DiT backbone for both arms, while selectively restricting information flow with an arm-mode attention mask. The input contains four arm-specific token groups:
\[
\begin{aligned}
\mathbf{X}_{\mathrm{in}}
&=[\mathbf{X}_{\mathrm{state}}^L,\mathbf{X}_{\mathrm{state}}^R,\\
&\quad\mathbf{X}_{\mathrm{action}}^L,\mathbf{X}_{\mathrm{action}}^R]\\
&\in\mathbb{R}^{N_{\mathrm{tok}}\times d},
\end{aligned}
\]
where $\mathbf{X}_{\mathrm{state}}^k$ denotes the state tokens of arm $k$ and $\mathbf{X}_{\mathrm{action}}^k$ denotes its noisy action tokens, with $k\in\{L,R\}$. For each execution mode $m\in\{\mathrm{left},\mathrm{right},\mathrm{dual}\}$, we define a binary mask
\[
\mathbf{M}^{(m)}\in\{0,1\}^{N_{\mathrm{tok}}\times N_{\mathrm{tok}}},
\]
and apply the mask to DiT self-attention:
\[
\begin{aligned}
&\operatorname{Attn}(\mathbf{Q},\mathbf{K},\mathbf{V}\mid\mathbf{M}^{(m)})\\
&\quad=\operatorname{softmax}\left(
\frac{\mathbf{Q}\mathbf{K}^\top}{\sqrt{d}}+\log\mathbf{M}^{(m)}
\right)\mathbf{V},
\end{aligned}
\]
where a zero entry in $\mathbf{M}^{(m)}$ suppresses the corresponding attention logit. In left-arm mode, only the left state and action groups can attend to one another; right-arm mode is defined symmetrically; dual-arm mode permits attention across all groups. The multimodal task representation remains shared across modes.

Dual-Mask uses the same flow-matching objective as the other variants. Its mask limits direct exchange between arm-specific token groups, but the two arms still share the DiT parameters and conditioning pathway. Dual-Tower therefore provides a stronger form of arm-level separation.

\subsubsection{Dual-Tower: Strong Decoupling via Independent Action Towers}
Dual-Tower uses two branches with the same architecture but independent parameters, denoted by $\mathcal{T}_L$ and $\mathcal{T}_R$. Each branch has its own state encoder, action tokenizer, DiT, and action decoder. Specifically, let $\mathbf{s}^k$ denote the raw proprioceptive state of arm $k$. The state encoder and action tokenizer produce
\[
\begin{aligned}
\mathbf{X}_{\mathrm{state}}^k
&=\operatorname{Enc}_{\mathrm{state}}^k(\mathbf{s}^k),\\
\mathbf{X}_{\mathrm{action}}^k
&=\operatorname{Tok}_{\mathrm{action}}^k(\mathbf{a}_{\alpha}^k),
\end{aligned}
\]
where $\mathbf{a}_{\alpha}^k$ is the noisy or interpolated action sequence used by flow matching. Thus, $\mathbf{X}_{\mathrm{state}}^k$ and $\mathbf{X}_{\mathrm{action}}^k$ denote encoded token sequences rather than raw states or actions. Their concatenated inputs are
\[
\begin{aligned}
\mathbf{X}_{\mathrm{in}}^L
&=[\mathbf{X}_{\mathrm{state}}^L,\mathbf{X}_{\mathrm{action}}^L]
\in\mathbb{R}^{N_{\mathrm{tok}}^L\times d},\\
\mathbf{X}_{\mathrm{in}}^R
&=[\mathbf{X}_{\mathrm{state}}^R,\mathbf{X}_{\mathrm{action}}^R]
\in\mathbb{R}^{N_{\mathrm{tok}}^R\times d}.
\end{aligned}
\]
The two sequences are processed independently and conditioned on the routed multimodal context. Their arm-specific states, action tokens, and parameters remain separate. During training, the annotated routing vector $\mathbf{r}=[r_L,r_R]$ selects $\mathcal{T}_L$, $\mathcal{T}_R$, or both; during inference, it is replaced by DAIR's prediction $\hat{\mathbf{r}}$. Dual-Tower approximately doubles the parameters of the arm-level components, while the frozen VLM and DAIR remain shared.

Only the selected tower contributes during single-arm execution; both towers receive the shared task context during dual-arm execution.

\paragraph{Routing-Conditioned Action Generation}
During training, the annotated $\mathbf{r}$ selects the arm-specific trajectory slices and action targets. Each active tower encodes its state and noisy action tokens, conditions the DiT on the routed multimodal context and predicted task progress, and decodes an arm-specific action. Dual-arm samples supervise both towers at synchronized time steps; inactive towers receive no action supervision for single-arm samples. At inference, $\hat{\mathbf{r}}$ controls the same data flow.

\paragraph{Arm-Conditioned View Masking}
Let $\mathbf{V}=[\mathbf{V}_{\mathrm{ego}},\mathbf{V}_L,\mathbf{V}_R]$ denote the SigLIP tokens from the three views and $\mathbf{E}_{\mathrm{lang}}$ the language-token sequence. For action conditioning, inactive end-effector-view tokens are masked before entering the VLM:
\[
\widetilde{\mathbf{V}}(\mathbf{r})
=\operatorname{Mask}\!\left(\mathbf{V},\mathbf{r}\right),
\qquad
\widetilde{\mathbf{E}}_{\mathrm{VL}}(\mathbf{r})
=\operatorname{VLM}\!\left(\widetilde{\mathbf{V}}(\mathbf{r}),\mathbf{E}_{\mathrm{lang}}\right).
\]
For $\mathbf{r}=[1,0]$, the right end-effector view is masked; the right-arm case is symmetric. For $\mathbf{r}=[1,1]$, both views are retained. The egocentric view remains available in all modes. Thus, the inactive end-effector view does not participate in the multimodal fusion used for action conditioning.

\paragraph{Active-Arm Supervision}
For Dual-Tower, the routing vector $\mathbf{r}=[r_L,r_R]$ selects the supervised towers:
\[
\mathcal{L}_{\mathrm{act}}
=r_L\mathcal{L}_{\mathrm{FM}}^L
+r_R\mathcal{L}_{\mathrm{FM}}^R.
\]
Here, $\mathcal{L}_{\mathrm{FM}}^k$ denotes the flow-matching objective for arm $k$. For single-arm samples, only one term is active, whereas both terms are used for dual-arm samples. Single-Tower and Dual-Mask use the same objective with their shared action-generation structures.

\subsection{Task-Progress Prediction Module}

Progress-aware control has been explored for understanding execution state and long-horizon subtask transitions~\cite{zhang2026vlac,ma2026furniturevla}. In TAO-DA, a GRU encodes the DiT temporal features $\mathbf{H}=[\mathbf{h}_1,\ldots,\mathbf{h}_T]$, and cross-attention fuses the resulting sequence with the VLM representation:
\[
\begin{aligned}
\mathbf{S}&=\operatorname{GRU}(\mathbf{H}),\\
\mathbf{Z}&=\operatorname{MHA}(\mathbf{S},\mathbf{E}_{\mathrm{VL}},\mathbf{E}_{\mathrm{VL}}).
\end{aligned}
\]
A regression head predicts $\hat{p}_t$ using normalized trajectory time as the target. At inference, this estimate supports DAIR's hold mechanism and the deployment-level scheduler.

\section{EXPERIMENTS}

\subsection{Platform and Dataset}

We collected 9,080 real-world teleoperation trajectories on an AGIBOT G1 humanoid robot with two 7-DoF arms and three RGB-D cameras operating at 30~Hz. Table~\ref{tab:dataset} summarizes the task composition. For routing evaluation, we randomly hold out 10\% of the data at the trajectory level; all routing variants use the same split. All action-policy variants use the same training data, observation interface, and action-execution strategy, while object placements vary across trials~\cite{zhaoVLARAILRealTimeAsynchronous2025}.

\begin{table}[tb]
\caption{Real-world teleoperation dataset.}
\label{tab:dataset}
\centering
\tablebody
\begin{tabular}{@{}llccc@{}}
\toprule
Category & Task & Mode & Traj. & Duration (h) \\
\midrule
\multirow{4}{*}{Single-arm} & Steamer Delivery & Single & 1,200 & 3.0 \\
& Bottle Grasping & Single & 880 & 3.2 \\
& Fruit Grasping & Single & 4,000 & 14.4 \\
& Tea Pouring & Single & 1,600 & 6.0 \\
\midrule
\multirow{2}{*}{Dual-arm} & Collaborative Tea & Dual & 500 & 2.0 \\
& Desktop Organization & Dual & 900 & 3.3 \\
\midrule
\textbf{Total} & -- & -- & \textbf{9,080} & \textbf{31.9} \\
\bottomrule
\end{tabular}
\end{table}

Single-arm tasks use one arm throughout execution, whereas dual-arm tasks involve simultaneous or sequential participation of both arms. Unless a task-specific protocol specifies otherwise, each real-robot condition is evaluated over 20 trials. Success is defined by the task goal: grasping the target object, placing it at the designated location, or completing tea pouring without spillage.

\subsection{Real-World Task Performance}

Tables~\ref{tab:single_arm_results} and~\ref{tab:dual_arm_results} report success rates on single- and dual-arm tasks. Each baseline uses its corresponding pretrained checkpoint and is fine-tuned on the same training data with the same training strategy and input configuration. For dual-arm evaluation, we include GR00T N1.5 and $\pi_{0.5}$, two representative baselines also evaluated on the complete single-arm task suite. A dash denotes a task that a model cannot execute.

\begin{table}[!t]
\caption{Single-arm task success rates (\%). The average includes only models evaluated on all five tasks.}
\label{tab:single_arm_results}
{\centering
\tablebody
\begin{tabular}{@{}lcccccc@{}}
\toprule
Task & GR00T & $\pi_0$ & $\pi_{0.5}$ & smolVLA & GO-1 & \textbf{TAO-DA} \\
\midrule
Bottle & 85 & 92.5 & \textbf{95} & 45 & 30 & \textbf{95} \\
Steamer-L & 90 & -- & 90 & -- & 57.5 & \textbf{94} \\
Steamer-R & \textbf{97.5} & -- & 75 & -- & 45 & 95 \\
Green Tea-L & 66.7 & -- & \textbf{75} & -- & 12.5 & \textbf{75} \\
Black Tea-R & 70.8 & -- & 67.5 & -- & 29.2 & \textbf{83} \\
Average & 82.0 & -- & 80.5 & -- & 34.8 & \textbf{88.4} \\
\bottomrule
\end{tabular}\par}

\vspace{0.5\baselineskip}

\caption{Success rates (\%) on real-world tasks requiring coordinated use of both arms.}
\label{tab:dual_arm_results}
{\centering
\tablebody
\begin{tabular}{@{}lcc@{}}
\toprule
Model & Collaborative Tea & Desktop Organization \\
\midrule
GR00T N1.5 & 78.5 & 80 \\
$\pi_{0.5}$ & \textbf{100} & 85 \\
\textbf{TAO-DA} & \textbf{100} & \textbf{95} \\
\bottomrule
\end{tabular}\par}
\end{table}

Among models evaluated on the complete single-arm suite, TAO-DA has the highest observed average success rate, 88.4\%, compared with 82.0\% for GR00T N1.5 and 80.5\% for $\pi_{0.5}$. It records the best or tied-best value on four tasks, while GR00T N1.5 is higher on Steamer Delivery-R. On the dual-arm tasks, TAO-DA records 100\% on Collaborative Tea and 95\% on Desktop Organization, matching the highest observed value on the former and exceeding both reported baseline values on the latter. These comparisons describe the observed outcomes under the corresponding task protocols and are not treated as statistical significance claims.

\begin{figure*}[tb]
    \centering
    \includegraphics[width=.9\linewidth,trim=0 70 0 50,clip]{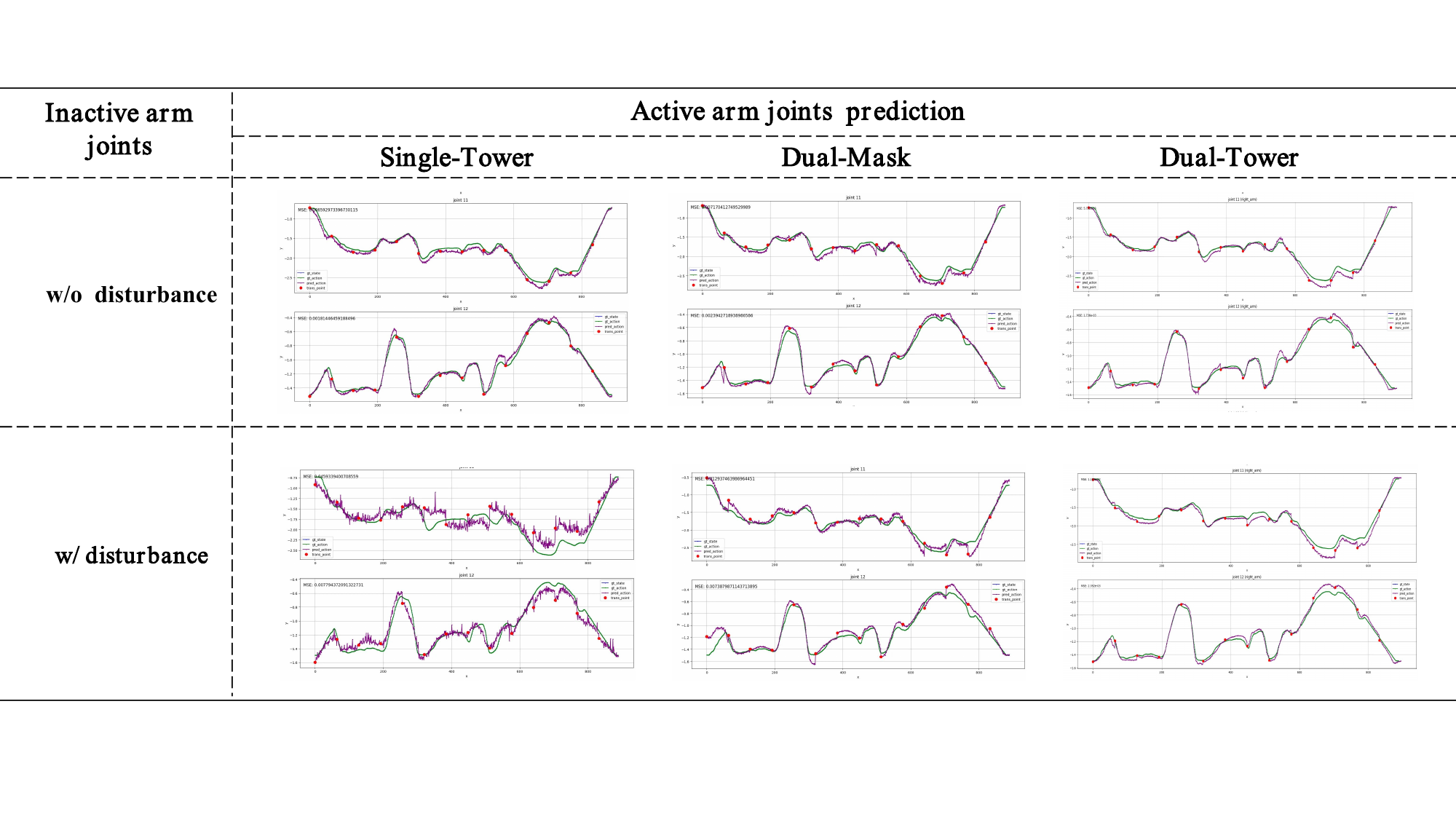}
    \caption{Active-arm trajectory predictions when the inactive arm is perturbed.}
    \label{fig:vis_wo_noise}
\end{figure*}

\subsection{Robustness to Inactive-Arm Disturbances}

We compare Single-Tower, Dual-Mask, and Dual-Tower during single-arm execution while perturbing the inactive arm. MSE is computed between offline model predictions and the ground-truth active-arm actions in the recorded trajectories under clean and disturbed conditions. The ratio is defined as $\mathrm{MSE}_{\mathrm{disturbance}}/\mathrm{MSE}_{\mathrm{clean}}$; a ratio near one indicates lower sensitivity to the disturbance.

\begin{table}[tb]
\caption{Offline robustness to inactive-arm disturbances.}
\label{tab:offline_interference}
\centering
\tablebody
\begin{tabular}{@{}llccc@{}}
\toprule
Active arm & Model & Clean MSE & Disturbed MSE & Ratio \\
\midrule
\multirow{3}{*}{Right} & Single-Tower & $2.69\times10^{-2}$ & $5.23\times10^{-2}$ & 1.94 \\
& Dual-Mask & $2.54\times10^{-2}$ & $3.73\times10^{-2}$ & 1.46 \\
& Dual-Tower & $2.83\times10^{-2}$ & $2.80\times10^{-2}$ & \textbf{0.99} \\
\midrule
\multirow{3}{*}{Left} & Single-Tower & $5.53\times10^{-2}$ & $8.85\times10^{-2}$ & 1.60 \\
& Dual-Mask & $5.88\times10^{-2}$ & $6.67\times10^{-2}$ & 1.13 \\
& Dual-Tower & $8.54\times10^{-2}$ & $8.54\times10^{-2}$ & \textbf{1.00} \\
\bottomrule
\end{tabular}
\end{table}

As shown in Table~\ref{tab:offline_interference}, the disturbed-to-clean MSE ratios for Single-Tower are 1.94 and 1.60 under right- and left-arm execution, respectively. Dual-Mask reduces these ratios to 1.46 and 1.13, whereas Dual-Tower yields ratios of 0.99 and 1.00. These results are consistent with the hypothesis that separate action towers reduce the sensitivity of the active arm to inactive-arm disturbances. Fig.~\ref{fig:vis_wo_noise} visualizes the corresponding active-arm trajectory predictions. Absolute clean-condition MSE should be interpreted separately from disturbance isolation.

We further evaluate real-robot execution from random initial states and under persistent inactive-arm disturbances. For the latter condition, noise is added to the seven-dimensional joint-state input of the inactive arm throughout task execution, while the active-arm state remains unchanged. Table~\ref{tab:realworld_interference} reports the resulting success rates. Across the random-initial-state and persistent-disturbance conditions, Dual-Tower maintains 95--100\% success across both active arms, whereas the corresponding ranges are 20--60\% for Single-Tower and 65--85\% for Dual-Mask.

\begin{table}[tb]
\caption{Real-robot success rates (\%) under the nominal condition, random initial states, and persistent inactive-arm disturbances (20 trials per condition).}
\label{tab:realworld_interference}
\centering
\tablebody
\begin{tabular}{@{}llccc@{}}
\toprule
Condition & \shortstack{Active\\arm} & \shortstack{Single-\\Tower} & \shortstack{Dual-\\Mask} & \shortstack{Dual-\\Tower} \\
\midrule
\multirow{2}{*}{Nominal} & Left & 80 & 90 & \textbf{95} \\
& Right & 70 & 95 & \textbf{100} \\
\midrule
\multirow{2}{*}{Random initial state} & Left & 40 & 85 & \textbf{95} \\
& Right & 60 & 80 & \textbf{95} \\
\midrule
\multirow{2}{*}{Persistent disturbance} & Left & 25 & 65 & \textbf{100} \\
& Right & 20 & 65 & \textbf{95} \\
\bottomrule
\end{tabular}
\end{table}

\begin{figure*}[tb]
    \centering
    \includegraphics[width=.9\linewidth]{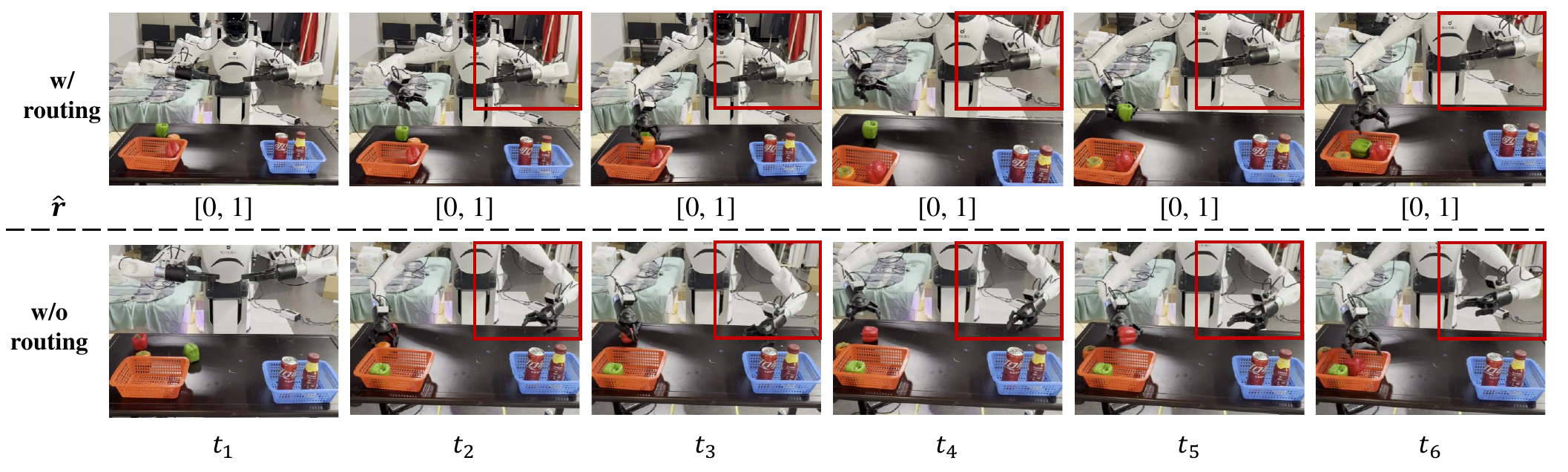}
    \caption{Qualitative comparison on Desktop Organization from identical initial conditions. The objects are located in the robot's right-arm workspace. Without arm-intent routing, the inactive left arm attempts an empty grasp; with DAIR, it remains stationary while the right arm executes the task.}
    \label{fig:routing_compare}
\end{figure*}

Under persistent disturbance, Dual-Tower records 100\% and 95\% success for left- and right-arm execution, compared with 25\% and 20\% for Single-Tower. The observed differences are 75 percentage points in both cases and are consistent with the offline isolation analysis. Given the finite trial count and the non-parameter-matched architectures, we interpret them as evidence within this protocol rather than as a general statistical claim.

\subsection{Arm-Intent Routing under Ambiguous Instructions}

We evaluate DAIR on tasks whose instructions do not explicitly identify the active-arm set. The classifier predicts left-only, right-only, or dual-arm intent. Fig.~\ref{fig:routing_compare} shows Desktop Organization from identical initial conditions, with objects only in the right-arm workspace. Without routing, the left arm attempts an empty grasp; with DAIR, it remains stationary while the right arm executes the task. This example illustrates the intended routing behavior but does not constitute a quantitative comparison of downstream task success.

On the held-out trajectory split, the full model with diverse instructions and task progress records per-class routing accuracies of 85.37\%, 83.31\%, and 96.13\% for left-only, right-only, and dual-arm intent, respectively.

\begin{figure}[!htb]
    \centering
    \includegraphics[width=.85\linewidth]{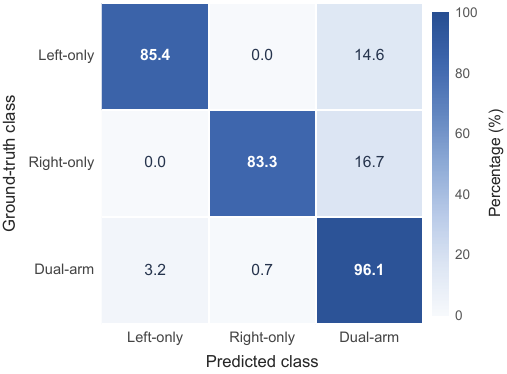}
    \caption{Row-normalized confusion matrix for left-only, right-only, and dual-arm intent classification. Each row reports percentages over the corresponding ground-truth class.}
    \label{fig:cm}
\end{figure}

The confusion matrix in Fig.~\ref{fig:cm} shows no cross-lateral errors between left-only and right-only predictions. Instead, 14.6\% of left-only samples and 16.7\% of right-only samples are misclassified as dual-arm. Thus, within this evaluation, errors are concentrated in distinguishing unimanual from bimanual execution rather than in identifying laterality.

\begin{figure}[!htb]
    \centering
    \includegraphics[width=\linewidth]{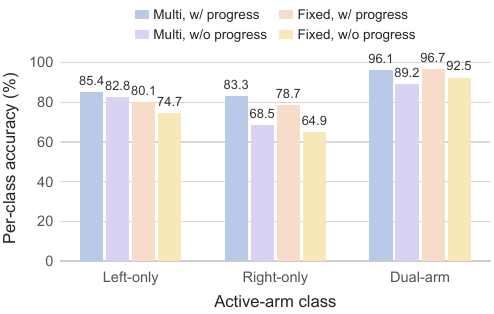}
    \caption{Per-class routing accuracy under instruction-diversity and task-progress ablations. Multi samples semantically equivalent instructions during training, whereas Fixed uses one instruction per task; w/o Progress removes the task-progress module.}
    \label{fig:routing_ac}
\end{figure}

\begin{figure*}[!t]
    \begin{minipage}[t]{0.48\linewidth}
        \centering
        \vspace{0pt}
        \includegraphics[width=\linewidth]{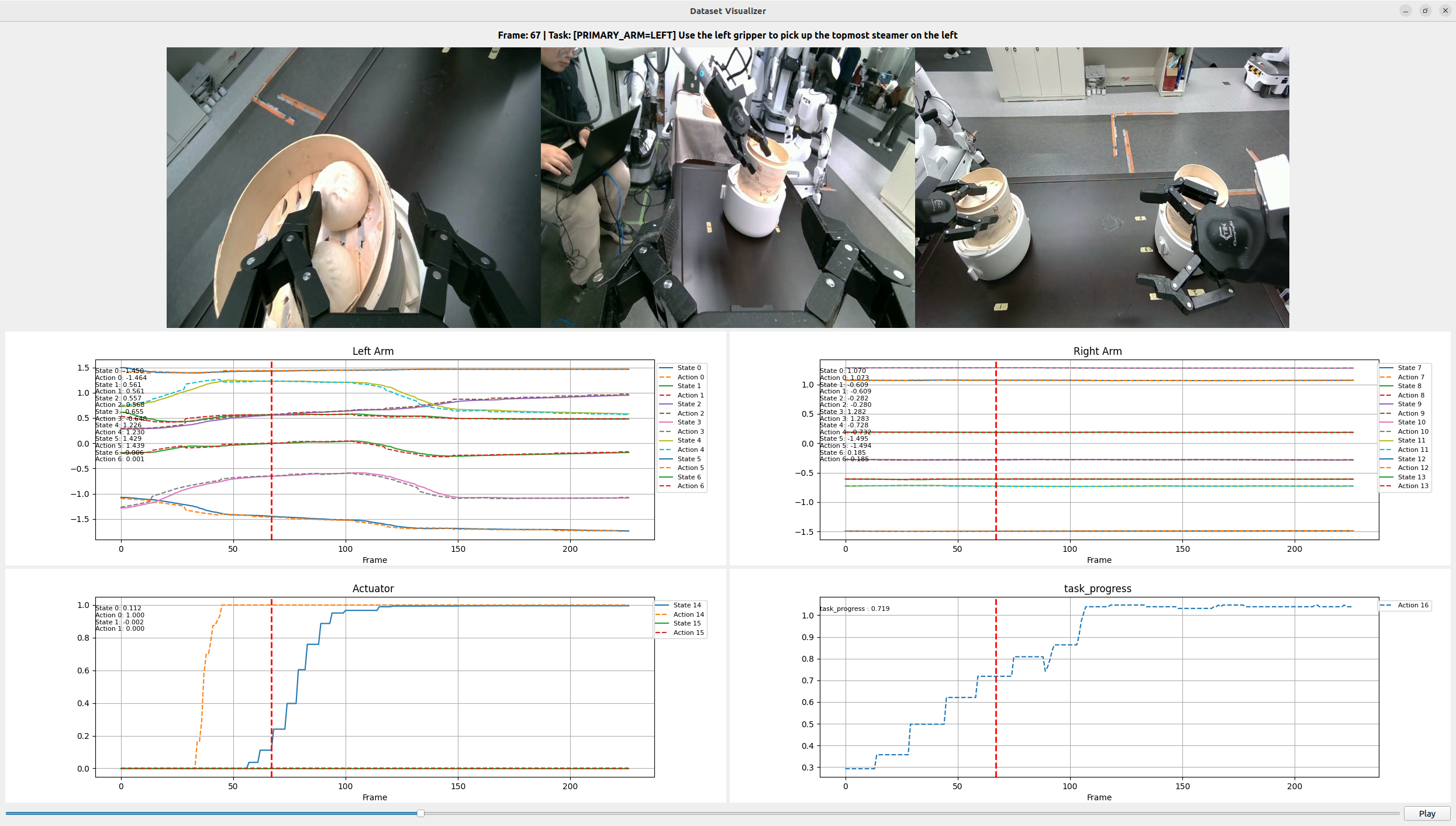}
        \caption{Qualitative visualization of task-progress prediction during real-world Steamer Delivery.}
        \label{fig:task_progress_vis}
    \end{minipage}
    \hfill
    \begin{minipage}[t]{0.48\linewidth}
        \centering
        \vspace{0pt}
        \includegraphics[width=.9\linewidth]{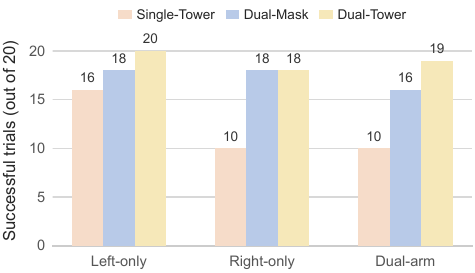}
        \caption{Number of successful Desktop Organization trials under left-only, right-only, and dual-arm execution. Each DAE variant is evaluated over 20 trials per execution mode and is trained on bimanual demonstrations without corresponding single-arm demonstrations.}
        \label{fig:dual_to_single}
    \end{minipage}
\end{figure*}

All four ablations in Fig.~\ref{fig:routing_ac} use the same held-out trajectories. Multi randomly samples a semantically equivalent instruction for each task during training, whereas Fixed uses one instruction per task during both training and evaluation; w/o Progress removes the entire task-progress module. Within Multi, removing progress changes right-only accuracy from 83.31\% to 68.48\% and dual-arm accuracy from 96.13\% to 89.22\%. With progress retained, Fixed records 78.66\% right-only accuracy, compared with 83.31\% for Multi. These protocol-specific differences characterize how routing accuracy changes with the availability of task-progress information and instruction diversity; they do not establish corresponding improvements in downstream task success. In addition, Fig.~\ref{fig:task_progress_vis} provides a qualitative visualization of the predicted task-progress signal during a real-world Steamer Delivery episode.

\begin{figure*}[!b]
    \centering
    \includegraphics[width=.9\linewidth]{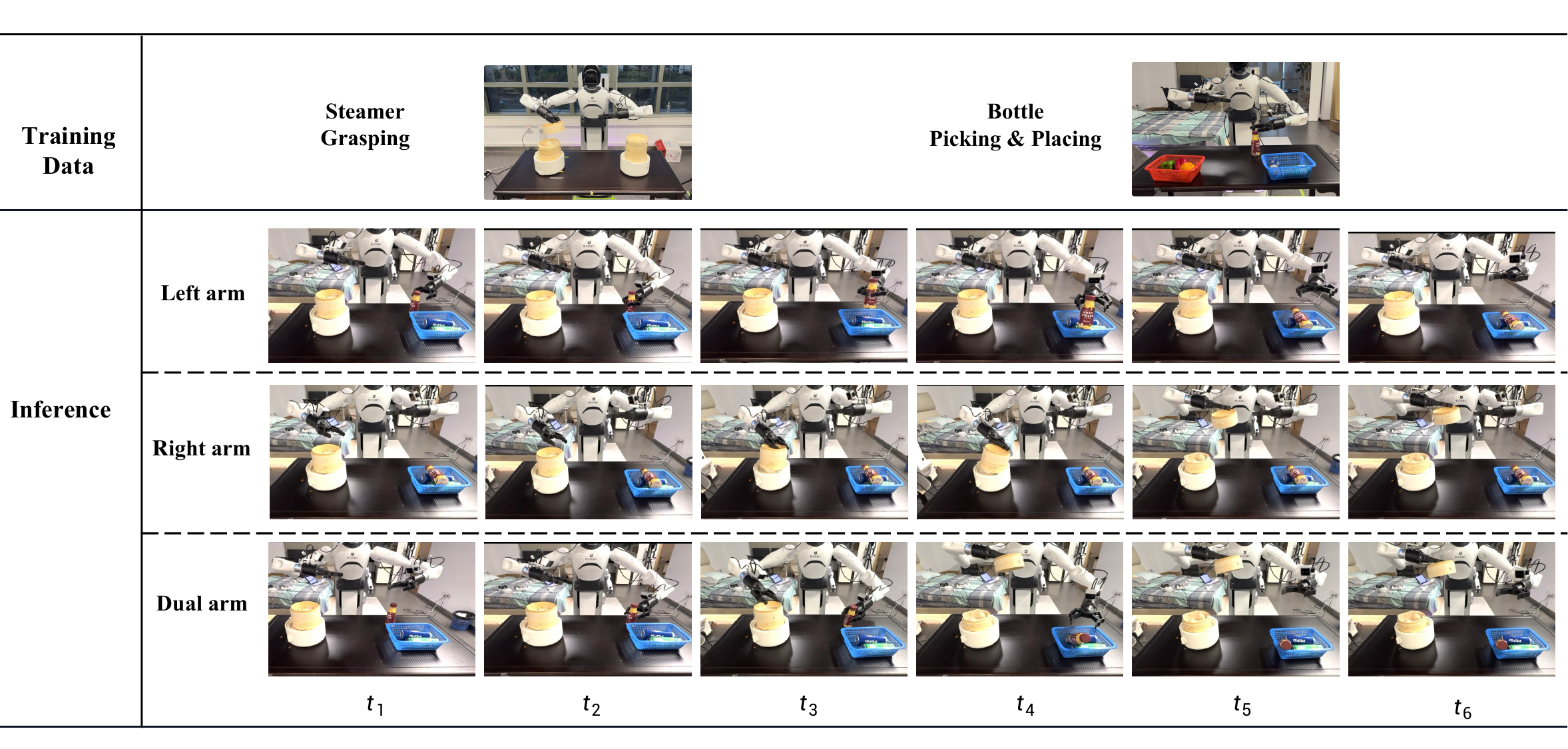}
    \caption{Qualitative single-to-dual-arm composition from separate arm-specific demonstrations. The two learned skills are executed in parallel without target-task bimanual demonstrations or additional fine-tuning.}
    \label{fig:skill_composition}
\end{figure*}

\subsection{Cross-Arm Skill Transfer and Composition}

We evaluate arm-wise skill reuse in three settings. First, all three DAE variants are trained on bimanual demonstrations of Collaborative Tea and Desktop Organization, without corresponding single-arm demonstrations, and evaluated on Desktop Organization under left-only, right-only, and dual-arm execution. As shown in Fig.~\ref{fig:dual_to_single}, the two decoupled variants record higher observed success rates than Single-Tower in all three modes over 20 trials per variant and mode. Dual-Tower reaches 100\%, 90\%, and 95\%, respectively, compared with 80\%, 50\%, and 50\% for Single-Tower. Under this protocol, the results are consistent with arm-specific pathways supporting reuse of behaviors learned from bimanual demonstrations.

Second, we train one Dual-Tower model on right-arm Steamer Grasping and left-arm Bottle Grasping. Fig.~\ref{fig:skill_composition} shows the model executing the two learned behaviors in parallel during bimanual Picking \& Placing, without bimanual demonstrations of the target task or additional fine-tuning. This is a qualitative instance of composing separately demonstrated arm-specific behaviors, rather than evidence of acquiring a new skill or tightly coupled coordination.

Finally, in three exploratory Collaborative Tea runs, we swap the training assignment of right-arm teapot grasping and left-arm cup grasping. Dual-Tower executes the reversed roles without reversed-role demonstrations or fine-tuning, but with lower qualitative precision; this limited observation is not treated as evidence of robust cross-arm transfer.

\subsection{Deployment Case Study}

\begin{figure}[!t]
    \centering
    \includegraphics[width=\linewidth]{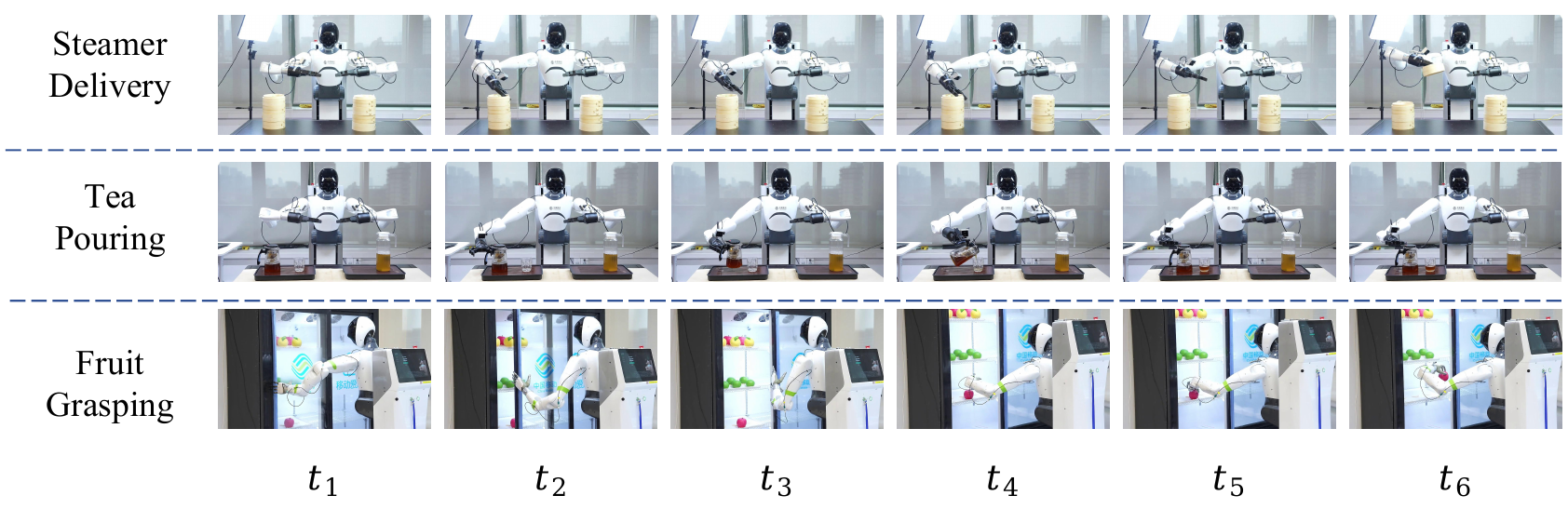}
    \caption{Representative tasks in the autonomous restaurant deployment.}
    \label{fig:restaurant_tasks}
\end{figure}

We also evaluate TAO-DA within an autonomous-restaurant deployment that integrates three robot types and coordinates six robots across more than 20 service tasks; representative tasks are shown in Fig.~\ref{fig:restaurant_tasks}. The complete system records 92.08\% end-to-end success over 101 full-process trials. Because this metric includes perception, manipulation, scheduling, and system execution, it characterizes the integrated deployment and is not attributed solely to TAO-DA. The task-progress signal supports both DAIR's hold mechanism and deployment-level task transitions.

\FloatBarrier
\subsection{Scope and Limitations}

Our conclusions are limited to the reported tasks and protocols. The architectural comparison is not parameter matched because Dual-Tower duplicates the arm-level components. DAIR is evaluated primarily through routing accuracy, and its downstream effect is illustrated qualitatively rather than isolated by task-success ablations. The disturbance study perturbs inactive-arm proprioception and does not cover all physical or visual disturbances. Routing uses the shared unmasked multimodal representation, whereas inactive wrist-view tokens are excluded from the VLM input used for action conditioning. Task progress is supervised by normalized trajectory time rather than semantic completion, and the composition and role-reversal studies remain qualitative.

\section{CONCLUSION}

We presented TAO-DA, which combines shared task-level multimodal reasoning with arm-specific action generation. Under the reported controlled disturbances, Dual-Tower reduces the observed sensitivity of active-arm execution to inactive-arm state perturbations, while DAIR provides explicit active-arm-set selection. The routing, skill-reuse, and deployment results characterize the framework within the evaluated protocols; broader validation will require parameter-matched comparisons, physical and visual disturbances, and quantitative transfer studies.

\bibliographystyle{IEEEbib}
\IEEEtriggeratref{5}
\bibliography{refs}

\end{document}